\documentclass[11pt]{article}

\usepackage[preprint]{acl}

\usepackage{times}
\usepackage{latexsym}
\usepackage{pifont}   
\usepackage{amssymb}    
 
\newcommand{\xmark}{\textcolor{red}{\ding{55}}}

\usepackage[T1]{fontenc}
\usepackage{tabularx}

\usepackage[utf8]{inputenc}

\usepackage{microtype}

\usepackage{inconsolata}
 \usepackage{amsmath}
\usepackage{graphicx}
\usepackage{tabularx}
\usepackage{hyperref}
\usepackage{url}
\usepackage{booktabs}
\usepackage{multirow}  
\usepackage{siunitx}
\usepackage{tcolorbox}
\usepackage[table]{xcolor}
\usepackage{float}
\usepackage{arydshln}
\usepackage{tikz}
\usetikzlibrary{positioning, arrows.meta}
\newcommand{\best}[1]{\textbf{#1}} 

\definecolor{rowgray}{RGB}{240,240,240} 
\usepackage{graphicx}

\title{SwarmSys: Decentralized Swarm-Inspired Agents for Scalable and Adaptive Reasoning}

\author{
\textbf{Ruohao Li\textsuperscript{*1}},
\textbf{Hongjun Liu\textsuperscript{*2,4}},
\textbf{Leyi Zhao\textsuperscript{5}},
\textbf{Zisu Li\textsuperscript{3}},
\textbf{Jiawei Li\textsuperscript{1}},\\
\textbf{Jiajun Jiang\textsuperscript{1}},
\textbf{Linning Xu\textsuperscript{6}},
\textbf{Chen Zhao\textsuperscript{2,4}},
\textbf{Mingming Fan\textsuperscript{\dag1,3}},
\textbf{Chen Liang\textsuperscript{\dag1}}\\
\textsuperscript{1}The Hong Kong University of Science and Technology (Guangzhou)\
\textsuperscript{2}New York University\\
\textsuperscript{3}The Hong Kong University of Science and Technology\
\textsuperscript{4}NYU Shanghai\\
\textsuperscript{5}Indiana University\
\textsuperscript{6}The Chinese University of Hong Kong\\
\small\textsuperscript{*}Equal contribution.
\small\textsuperscript{\dag}Co-corresponding authors.\\[2pt]
\small\textbf{Correspondence:}
\href{mailto:rli777@connect.hkust-gz.edu.cn}{rli777@connect.hkust-gz.edu.cn},
\href{mailto:lh3862@nyu.edu}{lh3862@nyu.edu}
}

\begin{document}
\maketitle
\begin{abstract}
Large language model (LLM) agents have shown remarkable reasoning abilities. However, existing multi-agent frameworks often rely on fixed roles or centralized control, limiting scalability and adaptability in long-horizon reasoning.
We introduce SwarmSys, a closed-loop framework for distributed multi-agent reasoning inspired by swarm intelligence. Coordination in SwarmSys emerges through iterative interactions among three specialized roles, Explorers, Workers, and Validators, that continuously cycle through exploration, exploitation, and validation.
To enable scalable and adaptive collaboration, we integrate adaptive agent and event profiles, embedding-based probabilistic matching, and a pheromone-inspired reinforcement mechanism, supporting dynamic task allocation and self-organizing convergence without global supervision.
Across symbolic reasoning, research synthesis, and scientific programming tasks, SwarmSys consistently outperforms baselines, improving both accuracy and reasoning stability. These findings highlight swarm-inspired coordination as a promising paradigm for scalable, robust, and adaptive multi-agent reasoning, suggesting that coordination scaling may rival model scaling in advancing LLM intelligence.
\end{abstract}


\section{Introduction}

The strong reasoning and planning capabilities of Large Language Models (LLMs) have spurred interest in multi-agent systems. These systems use collaboration to enhance reasoning diversity and reliability. Frameworks such as AutoGen~\cite{wu2023autogenenablingnextgenllm}  enable agent-to-agent dialogue for multi-agent applications, while CAMEL~\cite{li2023camelcommunicativeagentsmind} leverages role-playing and inception prompting to facilitate autonomous cooperation. AutoGen Studio further provides a no-code interface for designing and debugging agent workflows~\cite{dibia2024autogenstudionocodedeveloper}. However, most systems use fixed roles and static communication, which limits their adaptability. This rigidity leads to redundant exploration and inefficiency, especially in long-horizon or dynamic tasks.
\begin{figure}
    \centering
    \includegraphics[width=1.0\linewidth]{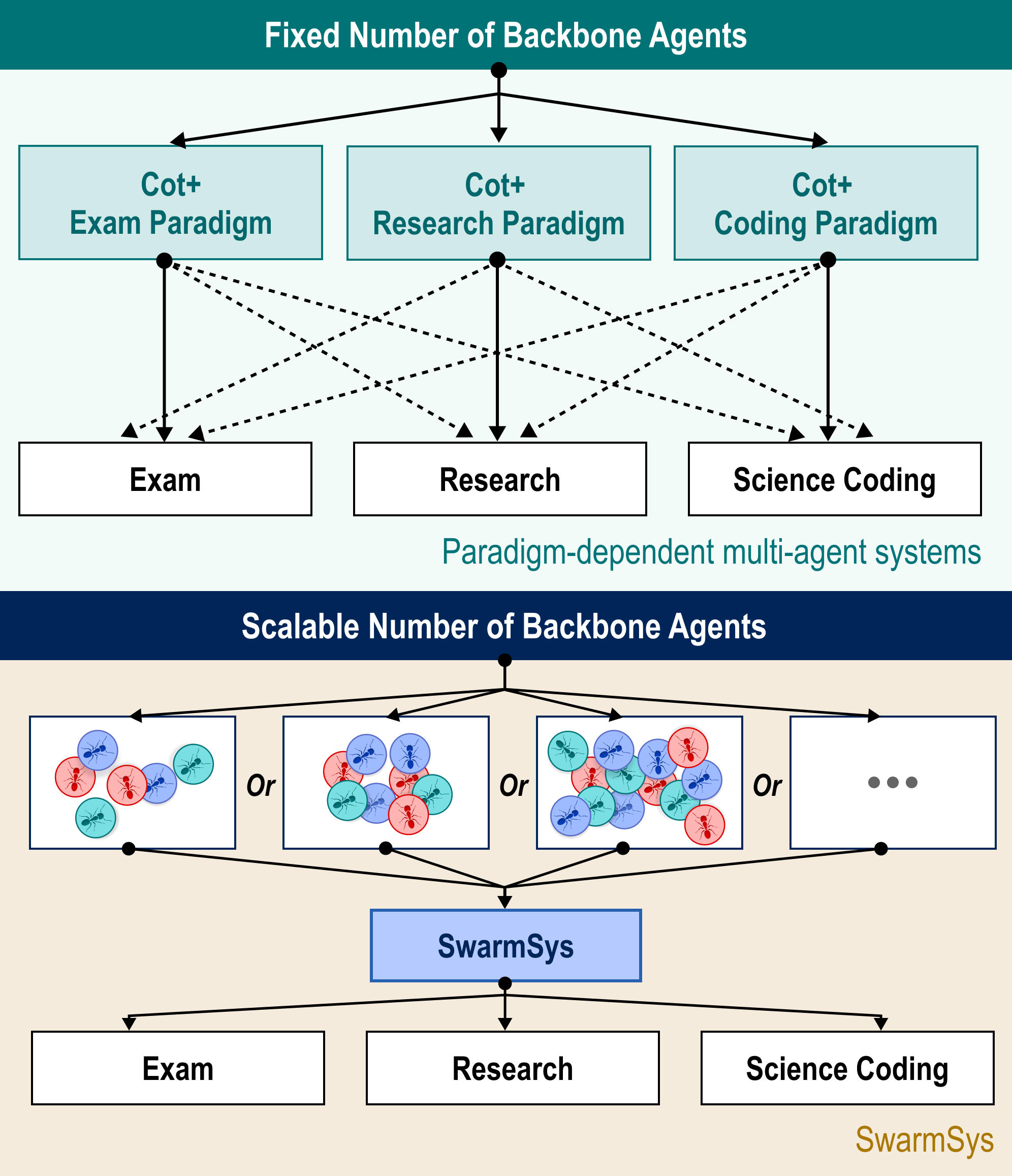}
    \caption{Comparison between paradigm-dependent multi-agent systems and SwarmSys. While existing methods rely on fixed, domain-specific agent paradigms, SwarmSys achieves scalable self-organization and cross-domain adaptability.}
    \label{fig1}
\end{figure}

\begin{figure*}[htbp]
    \centering
    \includegraphics[width=1.0\linewidth]{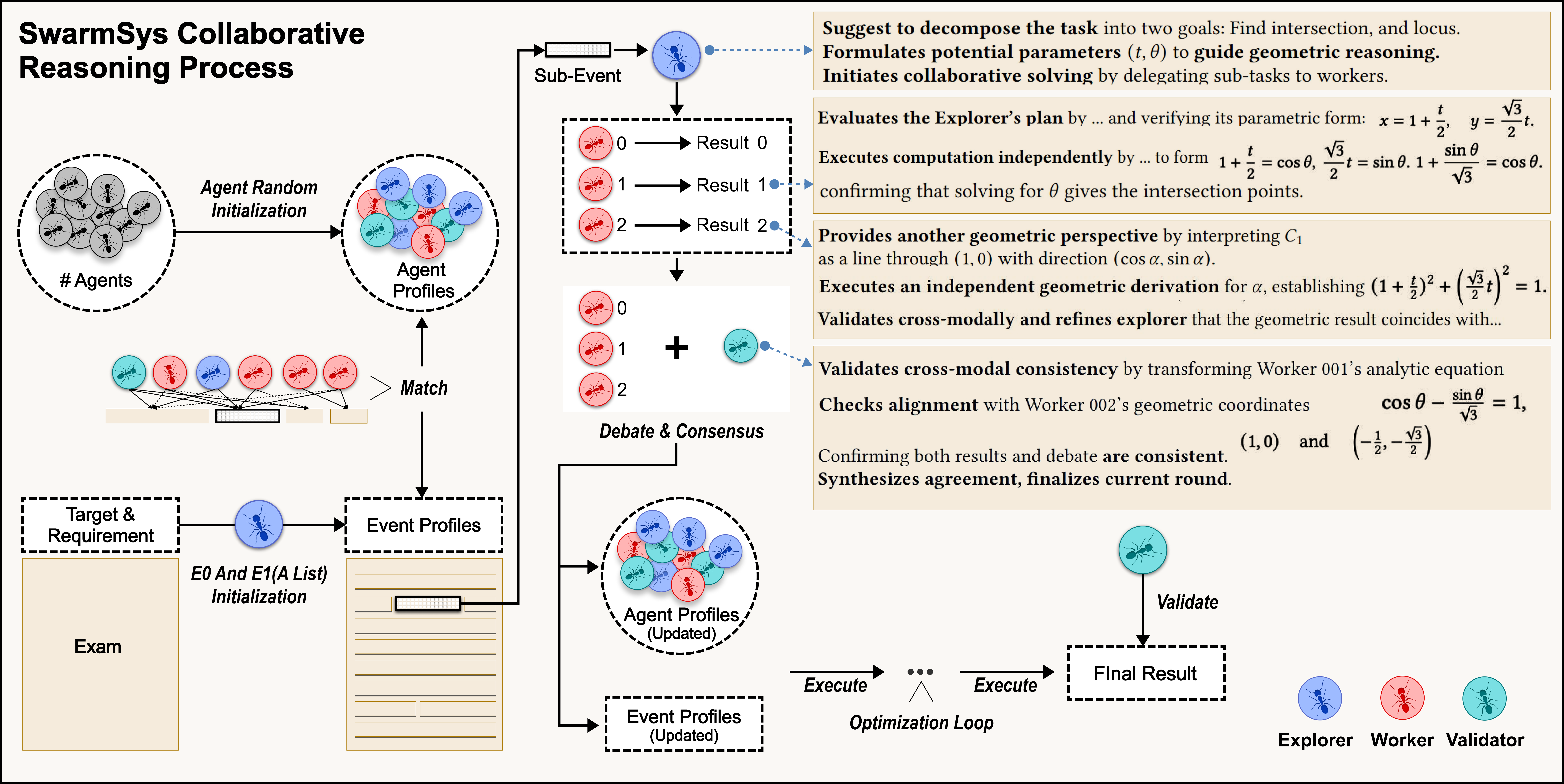}
    \caption{Overall workflow of the SwarmSys collaborative reasoning process. Each task is decomposed into sub-events handled by specialized agents: \textit{Explorers} propose solution paths, \textit{Workers} execute subtasks, and \textit{Validators} ensure consistency. Agents iteratively perform debate--consensus cycles that update event profiles and reinforce effective reasoning strategies until convergence.}
    \vspace{-1.5em}
    \label{fig:workflow}
\end{figure*}

To address limitations, recent work has explored adaptive or self-improving agent collaboration. AutoAgent enables natural-language-defined behaviors instead of hand-coded roles~\cite{tang2025autoagentfullyautomatedzerocodeframework}, while Mixture-of-Agents introduces layered coordination for more robust reasoning~\cite{wang2024mixtureofagentsenhanceslargelanguage}. However, these systems still rely on centralized orchestration or manually designed topologies, limiting scalability and long-term stability. Inspired by the decentralized intelligence of natural swarms, where simple signals regulate cooperation and task allocation~\cite{bonabeau1999swarm,dorigo2004ant}, we propose SwarmSys, a closed-loop framework that enables LLM agents to coordinate through lightweight, pheromone-like traces encoding contextual utility. This mechanism fosters self-organized collaboration, dynamically balancing exploration and convergence without centralized control.

Unlike debate or tree-based systems, SwarmSys allows coordination to emerge organically through iterative interaction and adaptive matching. Agents assume three roles, Explorers, Workers, and Validators, mirroring the division of labor in natural ant colonies. Explorers expand hypotheses, Workers refine and execute subtasks, and Validators ensure consistency, together forming continuous exploration–exploitation–validation cycles driving decentralized convergence. 
A core innovation is the use of profiles as adaptive memory. Agent and event profiles evolve with ability embeddings, workload, and context, enabling embedding-based reallocation and balanced participation—analogous to ants redistributing across foraging sites. Moreover, SwarmSys employs a pheromone-inspired reinforcement process: validated traces strengthen future compatibility, while ineffective ones decay, forming a decentralized optimization loop that enhances efficiency and stability over time.


Evaluated across symbolic reasoning, research synthesis, and scientific programming tasks, SwarmSys outperforms baselines such as GPTSwarm~\cite{zhuge2024languageagentsoptimizablegraphs}, achieving up to 10.7\% higher accuracy and 9.9\% better sub-task correctness. Remarkably, a swarm of GPT-4o-based agents approaches GPT-5 performance, showing that scaling coordination can substitute for model scaling. Qualitative analyses reveal emergent behaviors, knowledge diffusion, specialization balance, and self-regularization, hallmarks of collective intelligence.

In summary, our contributions are threefold:
(1) SwarmSys Framework: A closed-loop distributed multi-agent reasoning framework inspired by swarm intelligence for convergence.
(2) Adaptive Coordination Mechanism: An embedding-based matching and pheromone-inspired reinforcement process enabling dynamic agent–event allocation, self-organized collaboration, and stable long-horizon reasoning.
(3) Comprehensive Evaluation: Extensive experiments across diverse reasoning tasks reveal consistent gains and emergent collective intelligence, showing that scaling coordination can rival scaling model capacity.

\section{Methodology}
\subsection{Overview of SwarmSys}
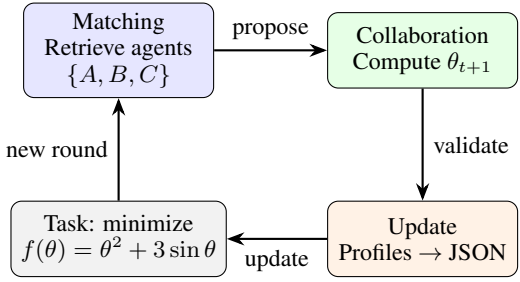
\begin{figure}[ht]
\centering
\begin{tikzpicture}[
    every node/.style={font=\small, align=center},
    box/.style={rounded corners, draw, minimum width=2.5cm, minimum height=1cm},
    >=Stealth
]
    
    \node[box, fill=blue!10] (matching) at (0,0) {Matching\\Retrieve agents\\$\{A, B, C\}$};
    \node[box, fill=green!10] (collab) at (4,0) {Collaboration\\Compute $\theta_{t+1}$};
    \node[box, fill=orange!10] (update) at (4,-2.5) {Update\\Profiles $\rightarrow$ JSON};
    \node[box, fill=gray!10] (task) at (0,-2.5) {Task: minimize\\$f(\theta)=\theta^2+3\sin\theta$};
    
    \draw[->, thick] (matching) -- node[above] {\footnotesize propose} (collab);
    \draw[->, thick] (collab) -- node[right] {\footnotesize validate} (update);
    \draw[->, thick] (update) -- node[below] {\footnotesize update} (task);
    \draw[->, thick] (task) -- node[left] {\footnotesize new round} (matching);
\end{tikzpicture}
\caption{Iterative cycle in SwarmSys: minimize $f(\theta)$ through matching → collaboration → update cycle.}
\label{fig:swarm_cycle}
\end{figure}
We present SwarmSys, a closed-loop collaborative framework for distributed problem solving. Unlike centralized orchestration or static task assignment, it converges through iterative matching → collaboration → update cycles, as is shown in 
Figure \ref{fig:swarm_cycle}. Upon receiving new task, event profiles are instantiated, and candidate agents are retrieved via embedding-based matching. Agents then enter collaborative rounds: explorers propose decomposition, workers execute subtasks after consensus, and validators verify intermediate results. After each round, both agent and event profiles are updated and stored for transparency and traceability. These evolving profiles feed subsequent iterations, enabling self-organization and adaptation. As shown in our framework (see figure \ref{fig:workflow}), through repeated cycles, we achieve high-quality solutions without central controller, relying instead on distributed collaboration and profile-driven adaptation. 

\subsection{Profiles as Adaptive Memory Units}
A core innovation of SwarmSys is the use of \textit{profiles} as adaptive memory units that record both context and accumulated experience.  
Each agent profile maintains identifiers, role information, ability descriptions, workload status, and longitudinal performance history, the profile format is provided in Appendix \ref{profile format}. 
Profiles evolve dynamically: embeddings and workload indicators are refreshed after each round, while historical success or failure influences future task matching. 
Agents thus behave as adaptive collaborators rather than stateless executors.  
Each event profile serves as a dynamic record of problem-solving. 
It contains task descriptions, dependency structures, metadata (e.g., composite or leaf type), progress logs, and participating agent lists. 
Over time, static specifications evolve into rich, evolving traces of reasoning.  

Together, these profiles constitute SwarmSys’s distributed memory: agent profiles encode competence and reliability, while event profiles track task evolution. 
Their joint updates ensure collaboration remains adaptive, interpretable, and transparent.





\subsection{Embedding-Based Matching with Exploration--Exploitation Dynamics}

The second key innovation is our embedding-based agent--event matching algorithm, which balances expertise, workload, and exploration.  

\paragraph{Agent embeddings.}
Each agent $A_i$ is represented by two embeddings capturing its ability and availability.  
The competence embedding is derived from the agent’s declared abilities and historical performance, processed through an instruction-guided encoder that contextualizes the agent’s prior experience for the current task. 
The availability embedding reflects workload and readiness, obtained from the agent’s status signals.  
These two vectors are summed to form the agent’s overall representation, combining long-term expertise with short-term availability cues, the the detailed process is provided in Appendix~\ref{appendix:agent_profile_and_behavior}.  
This design allows the system to favor agents who are both skilled and currently underutilized.

\paragraph{Event embeddings.}
Each event $E_j$ is encoded through an instruction-conditioned embedding that integrates its textual description, dependency relations, progress state, and milestone metadata, the method is listed at Appendix\ref{app:prompt_app}. 
This representation captures both the semantic meaning of the task and its structural role within the reasoning process. 
By embedding events and agents in a shared latent space, SwarmSys can estimate their compatibility in a continuous and scalable manner.

\paragraph{Compatibility and decision dynamics.}
The compatibility between an agent and an event is measured by normalized cosine similarity between their embeddings, ensuring a value between 0 and 1 for interpretability.  
To prevent premature convergence and encourage exploration, SwarmSys adopts a dynamic $\varepsilon$-greedy policy.  
Each agent explores new matches with probability $\varepsilon_i$ and exploits high-compatibility matches otherwise.  
The exploration rate $\varepsilon_i$ adapts to recent performance: agents with high average success explore less, while underperforming agents explore more.  
Empirically, we initialize $\varepsilon_i$ around 0.15 to maintain minimal randomness and allow it to fluctuate within a small range (up to 0.35) depending on recent success. The values are designed based on natural ant behaviors~\cite{Lecheval2024.02.20.581181}.
This ensures that exploration gradually decreases as the system stabilizes.  
During exploration, matches are sampled proportionally to similarity, enabling serendipitous but plausible pairings.  
During exploitation, a sigmoid-weighted sampling function emphasizes strong compatibility, controlled by a sharpness factor $\gamma$ that modulates selectivity. The detailed behavioral derivation process is provided in Appendix~\ref{appendix:agent_profile_and_behavior}. This mechanism enables three properties: adaptivity through evolving embeddings, stability through probabilistic sampling, and robustness by balancing exploration with exploitation.

\subsection{Pheromone-Inspired Optimization}
Finally, SwarmSys incorporates a pheromone-inspired optimization process to refine allocation and solution quality. Each validated contribution reinforces the compatibility between an agent and an event, updating embeddings and increasing the likelihood of similar matches in future rounds. Idle or invalid matches, by contrast, receive no reinforcement and gradually decline in competitiveness as other profiles evolve, mimicking pheromone evaporation without explicit decay.

This implicit reinforcement--evaporation dynamic complements the exploration--exploitation policy. Exploration guarantees diversity and prevents deadlock, exploitation prioritizes promising matches, and bounded probabilities ensure stability. As explorers, workers, and validators collaborate across rounds, SwarmSys converges to high-quality solutions while maintaining flexibility and resilience in search dynamics.

\section{Experiment}

\subsection{Experiment Setting}
We evaluate SwarmSys across three reasoning categories that collectively span symbolic computation, open-domain research synthesis, and scientific programming.
All evaluations use dataset-specific metrics following their official definitions to ensure fair comparison.

\paragraph{Baselines}
Since agents show domain-specific strengths, we select the strongest baseline for each task category instead of using a uniform set. 
For exam-style reasoning, we compare against GPT-4o-based IO (direct LLM invocation) \cite{openai2024gpt4ocard}, CoT \cite{10.5555/3600270.3602070}, CoT-SC \cite{wang2022self}, Self-Refine \cite{madaan2023selfrefineiterativerefinementselffeedback}, MultiPersona \cite{wang2023unleashing}, and GPTSwarm \cite{zhuge2024languageagentsoptimizablegraphs}.  
For research tasks, we include general-purpose baselines (IO, CoT, Self-Refine), and deep research agents (Grok Deeper Search, DeepResearchAgent).  
For scientific programming, we use both general-purpose reasoning systems (Self-Refine, CoT) and domain-specific agents (GPTSwarm, DeepResearchAgent).  
We also report results from GPT-5 as an upper bound for single-agent performance. 

\paragraph{Dataset}
Table~\ref{tab:dataset_stat} summarizes the four benchmarks, covering quantitative, analytical, and procedural reasoning. 
This diversity ensures that SwarmSys is evaluated across both discrete symbolic reasoning and open-ended research generation settings. More details of our dataset settings are shown in Appendix \ref{app:dataset_app}
\begin{table}[ht]
\centering
\caption{Overview of the four reasoning benchmarks used in our experiments. Each dataset differs in domain focus, reasoning type, and data format.} 
\label{tab:dataset_stat}
\resizebox{\columnwidth}{!}{
\begin{tabular}{lccc}
\toprule
\textbf{Dataset} & \textbf{Focus} & \textbf{Reasoning} & \textbf{\#Samples} \\
\midrule
\textbf{GaoKao Bench}~\cite{zhang2024evaluatingperformancelargelanguage} & Quantitative \& Cross-domain & Symbolic  & 800 \\
\textbf{Omni-Math}~\cite{gao2024omnimathuniversalolympiadlevel} & Hard-Level Quantitative & Conceptual  & 300 \\
\textbf{DeepResearch}~\cite{du2025deepresearchbenchcomprehensivebenchmark} & Scientific QA & Analytical  & 200 \\
\textbf{SciCode}~\cite{tian2024scicoderesearchcodingbenchmark} & Computational & Procedural & 338 \\
\bottomrule
\end{tabular}
}
\vspace{-1.5em}
\end{table}

\begin{table*}[ht]
\centering
\caption{Performance comparison on \textbf{Exam-style tasks} (single-, multi-subject, and Olympic-level).
Metrics: Accuracy (Acc.) and Knowledge Coverage (Cov.). SwarmSys-8 means the system contains 8 agents (1* explorer, 6*workers, 1*validator).}
\label{tab:exam-results}
\small
\begin{tabular}{l ccc ccc cc}
\toprule
\multirow{2}{*}{\textbf{Method}} &
\multicolumn{2}{c}{\textbf{Math Exam}} &
\multicolumn{2}{c}{\textbf{STEM Mix}} &
\multicolumn{2}{c}{\textbf{Olympic Math}} &
\multicolumn{2}{c}{\textbf{Average}} \\
\cmidrule(lr){2-3}\cmidrule(lr){4-5}\cmidrule(lr){6-7}\cmidrule(lr){8-9}
 & Acc. & Cov. & Acc. & Cov. & Acc. & Cov. & Acc. & Cov. \\
\midrule
IO (GPT-4o)                & 46.3 & 45.7 & 57.4 & 55.2 & 23.5 & 57.6 & 42.4 & 52.8 \\
CoT (GPT-4o)               & 52.6 & 49.0 & 60.8 & 59.7 & 30.0 & 66.3 & 47.8 & 58.3 \\
CoT-SC (5-shot)            & 63.2 & 62.4 & 64.7 & 62.8 & 28.3 & 51.7 & 52.1 & 59.0 \\
Self-Refine (GPT-4o)       & 79.3 & 9.8 & 70.5 & 79.0 & 16.6 & 45.0 & 55.5 & 44.6 \\
MultiPersona (GPT-4o)      & 52.4 & 51.4 & 60.9 & 62.3 & 35.0 & 68.3 & 49.4 & 60.7 \\
GPTSwarm\textsuperscript{$\dagger$} & 65.5 & 70.3 & 69.6 & 71.4 & 40.0 & 73.3 & 58.4 & 71.7 \\
GPT-5                      & 87.2 & 90.6 & 87.7 & 89.1 & 31.2 & 70.5 & 68.7 & 83.4 \\
\midrule
\rowcolor{rowgray}
\textbf{SwarmSys-8 (Ours)} & \textbf{76.2} & \textbf{80.2} & \textbf{78.7} & \textbf{81.3} & \textbf{42.4} & \textbf{73.2} & \textbf{65.8} & \textbf{78.2} \\
\bottomrule
\end{tabular}
\end{table*}

\begin{table*}[ht]
\centering
\caption{Performance comparison on \textbf{Research tasks} (DeepResearch Bench). 
Metrics: Comprehensiveness, Depth, Instruction Following, and Readability.}
\label{tab:research-results}
\small
\begin{tabular}{l cccc c}
\toprule
\textbf{Method} & \textbf{Overall} & \textbf{Comp.} & \textbf{Depth} & \textbf{Inst.} & \textbf{Read.} \\
\midrule
IO (GPT-4o-Search)     & 30.7 & 27.8 & 20.4 & 41.0 & 37.6 \\
CoT (GPT-4o)           & 29.3 & 29.5 & 22.8 & 33.5 & 36.8 \\
Self-Refine            & 35.9 & 35.4 & 27.0 & 44.1 & 41.0 \\
DeepResearchAgent      & 48.8 & 48.5 & 48.5 & 49.1 & 49.4 \\
Grok Deeper Search     & 40.2 & 37.9 & 35.3 & 46.3 & 44.0 \\
\midrule
\rowcolor{rowgray}
\textbf{SwarmSys-8 (Ours)} & \textbf{42.5} & \textbf{39.6} & \textbf{38.0} & \textbf{50.0} & \textbf{46.3} \\
\bottomrule
\end{tabular}
\vspace{-1.5em}
\end{table*}
\begin{table}[ht]
\centering
\caption{Performance comparison on \textbf{Science Coding tasks} (SciCode benchmark). 
Metrics: Pass@Main and Pass@Sub (percentages). $\dagger$ No longer available; metrics from SciCode report.} 
\label{tab:scicode-results}
\small
\begin{tabular}{lcc}
\toprule
\textbf{Method} & \textbf{Pass@Main} & \textbf{Pass@Sub} \\
\midrule
IO (GPT-4o)                        & 2.0  & 28.3 \\
OpenAI o3-mini-medium\textsuperscript{$\dagger$} & 9.2  & 34.4 \\
CoT-SC (GPT-4o)                    & 8.8  & 28.7 \\
Self-Refine                        & 10.0 & 33.3 \\
GPTSwarm                           & 8.6  & 29.2 \\
\midrule
\rowcolor{rowgray}
\textbf{SwarmSys-14 (Ours)}        & \textbf{12.5} & \textbf{45.2} \\
\bottomrule
\end{tabular}
\end{table}
\paragraph{Metries}
We evaluate SwarmSys on three reasoning categories, using the original domain-specific metrics and protocols from each dataset
(1) \textit{Exam-style reasoning} tasks (Math Exam, STEM Mix, and Olympic Math) use Accuracy and Knowledge Coverage as defined in prior benchmark releases. For Math Exam and STEM Mix, we use subset from GAOKAO Bench. For Olympic Math, we use subset from Omni-Math and rearranged them to comply with real-world exam.
(2) \textit{Research-level reasoning} tasks (DeepResearch Bench) employ composite metrics including RACE (Comprehensiveness, Depth, Instruction Following, Readability) and FACT (Citation Accuracy, Effective Citation).
(3) \textit{Science Coding} tasks (SciCode) are evaluated using Pass@Main and Pass@Sub metrics, capturing correctness at both task and sub-task levels.
All metrics follow their dataset definitions to ensure faithful comparison and reproducibility.

\begin{table*}[!t]
\centering
\caption{Ablation on Exam task under varying numbers of Agents ($A$). 
Metrics: Accuracy (Acc., \%), Knowledge Coverage (Cov., \%). }
\label{tab:ablation-workers}
\small
\begin{tabular}{
l
S[table-format=2.1] S[table-format=2.1]
S[table-format=2.1] S[table-format=2.1]
S[table-format=2.1] S[table-format=2.1]
S[table-format=2.1] S[table-format=2.1]
S[table-format=2.1] S[table-format=2.1]
}
\toprule
\multirow{2}{*}{\textbf{Method}} &
\multicolumn{2}{c}{$A{=}4$} &
\multicolumn{2}{c}{$A{=}8$} &
\multicolumn{2}{c}{$A{=}14$} &
\multicolumn{2}{c}{$A{=}20$} &
\multicolumn{2}{c}{$A{=}32$} \\
\cmidrule(lr){2-11}
 & {Acc.} & {Cov.}  
 & {Acc.} & {Cov.} 
 & {Acc.} & {Cov.} 
 & {Acc.} & {Cov.} 
 & {Acc.} & {Cov.} \\
\midrule
Rand-NoRoles & 43.2 & 41.4 & 42.9 & 42.1 & 44.1 & 43.4 & 44.3 & 43.2 & 43.8 & 42.3 \\
Rand-Roles   & 56.3 & 52.4 & 58.2 & 54.8 & 58.5 & 56.1 & 58.3 & 56.0 & 57.3 & 55.9 \\
\rowcolor{rowgray}
\textbf{SwarmSys} & \best{74.3} & \best{79.7} & \best{76.2} & \best{80.2} & \best{77.3} & \best{81.0} & \best{77.2} & \best{81.3} & \best{76.5} & \best{80.8} \\
\bottomrule
\end{tabular}
\end{table*}

\subsection{Overall Results}

\paragraph{Exam-style Reasoning.}
Table~\ref{tab:exam-results} shows that SwarmSys consistently outperforms all GPT-4o-based multi-agent baselines on both single- and multi-subject exams, achieving an average improvement of $+12.5\%$ Accuracy and $+10.8\%$ Coverage over GPTSwarm.  
While GPT-5 achieves the strongest absolute scores, SwarmSys-8 narrows the gap by over $70\%$, demonstrating that swarm-level cooperation can approach next-generation model performance without access to stronger backbones.  
Qualitatively, we observe that SwarmSys agents exhibit complementary specialization: Explorers diversify problem-solving strategies, while Validators efficiently prune redundant reasoning chains, leading to higher coverage with reduced variance.

\paragraph{Research-level Reasoning.}
As shown in Table~\ref{tab:research-results}, SwarmSys surpasses Grok Deeper Search in overall RACE score (+2.3\%) and instruction-following (+3.7\%), reflecting the benefit of distributed role specialization in literature synthesis and factual consolidation.  
SwarmSys achieves especially large gains in comprehensiveness and readability, suggesting that swarm debates improve global coherence even in open-ended research generation tasks.

\begin{figure*}[ht]
    \centering
    \includegraphics[width=1\linewidth]{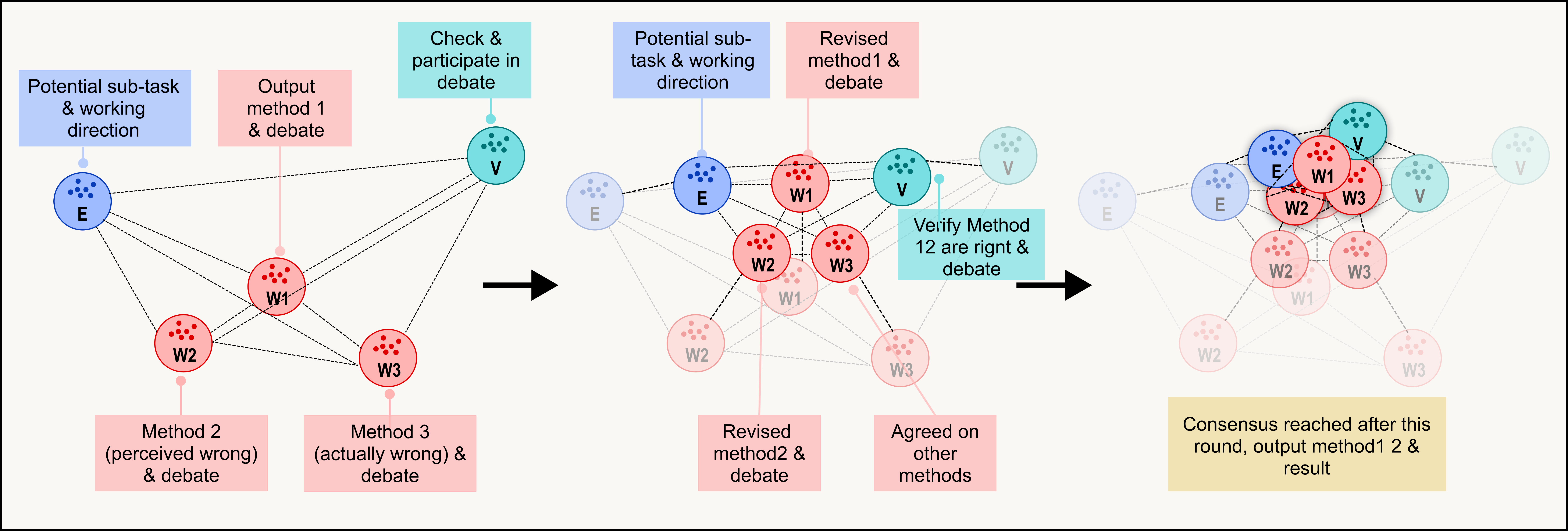}
    \caption{
    Swarm reasoning trajectory on MathExam.
    Explorers initiate sub-tasks, Workers debate and revise alternative methods, and Validators enforce cross-checks across rounds. 
    The swarm collectively converges to consistent solutions through debate-driven consensus formation.
    }
    \vspace{-1.5em}
    \label{fig:swarm-effect}
\end{figure*}

\paragraph{Scientific Programming.}
In Table~\ref{tab:scicode-results}, SwarmSys demonstrates notable improvements on SciCode: $+2.5\%$ Pass@Main and $+11.9\%$ Pass@Sub over the best GPT-4o baselines.  
These results indicate that dynamic role coordination and progressive refinement effectively decompose complex computational problems.  
Interestingly, Pass@Sub improvements are more pronounced, supporting our hypothesis that swarm collaboration benefits from modular code generation and localized validation.

\subsection{Ablation Study}
We ablate the design of SwarmSys along three axes: \textit{(a)} removing dynamic profile updates and embedding-based matching (Ours-Roles-Rand); \textit{(b)} removing both roles and matching, leaving homogeneous random assignment (Rand-NoRoles); and \textit{(c)} varying the number of Agents $A$ from $4$ to $32$ (Table~\ref{tab:ablation-workers}), while ensure that each event contains at least one explorer, one worker, and one validator.  
All variants use identical backbones and decoding parameters.
Results reveal three trends: (1) Removing roles leads to a substantial decline in both accuracy and coverage (e.g., $56.3\%\!\to\!43.2\%$ in accuracy and $52.4\%\!\to\!41.4\%$ in coverage at $A{=}4$), confirming that cooperative debate and functional specialization are essential for maintaining diversity without redundancy.
(2) Adaptive matching and profiling further enhance performance: embedding-based matching increases coverage by up to $+27.3\%$ by aligning agent capabilities with task semantics.  
(3) Scaling saturates around $W{=}14$: while both metrics improve with more Workers, gains plateau beyond $14$, indicating that agent saturation occurs once the subtask granularity is fully covered.  

\subsection{Swarm Effect: Emergent Collective Intelligence}

Our design philosophy centers on the Swarm Effect: a collection of properly coordinated, limited agents can collectively approximate or surpass a stronger single-agent model.
Figure~\ref{fig:swarm-effect} visualizes how SwarmSys (GPT-4o backbone) gradually approaches the performance of stronger models such as GPT-5 and DeepResearchAgent as the swarm size increases.

\paragraph{Emergent Performance Scaling.}
Quantitatively, as the number of active agents increases from $A{=}4$ to $A{=}14$, both Accuracy and Knowledge Coverage improve consistently 
(e.g., from $74.3/79.7$ to $77.3/81.0$ on the Exam task). 
However, further expansion to $A{=}20$ or $A{=}32$ yields negligible gains (within $<1\%$ difference), indicating that agent capacity saturates once subtask diversity and knowledge space have been sufficiently covered. 
This scaling curve mirrors real-world swarm systems, where the marginal utility of additional workers decreases after local niches become saturated.

\paragraph{Collaborative Dynamics.}
Unlike conventional multi-agent ensembles that rely on static voting, SwarmSys agents interact through pheromone-inspired event matching and debate-driven validation.
Explorers dynamically propose new sub-goals, Workers attempt partial solutions, and Validators consolidate outcomes based on collective memory.
This dynamic feedback loop creates a self-organizing division of labor where each agent adapts its behavior not from external commands but through the evolving swarm state.
Empirically, this leads to:
(i) higher diversity in reasoning paths, and 
(ii) smoother convergence across reasoning rounds.

\paragraph{Knowledge Diffusion and Self-Regularization.}
We further observe that intermediate reasoning traces in SwarmSys display emergent \textit{knowledge diffusion}: 
factual entities, equations, or hypotheses discovered by one agent are reused, revised, or even corrected by others without explicit synchronization.
This effect increases factual precision while maintaining interpretability.
In qualitative analyses, the system exhibits an implicit \textit{regularization behavior}—weaker agents’ errors are diluted by consensus mechanisms, preventing local hallucinations from dominating global output.

\paragraph{From Coordination to Intelligence.}
The Swarm Effect demonstrates that collective intelligence is not a linear function of model size, 
but an emergent property of structured interaction. 
While individual agents are limited by GPT-4o, the swarm collectively builds a distributed memory and decision space, enabling generalization beyond any single agent.
This property highlights a promising direction for future large-scale reasoning systems: 
scaling through coordination rather than parameter count.

\section{Qualitative Analysis}
\begin{figure}[t]
\centering
    \includegraphics[width=\linewidth]{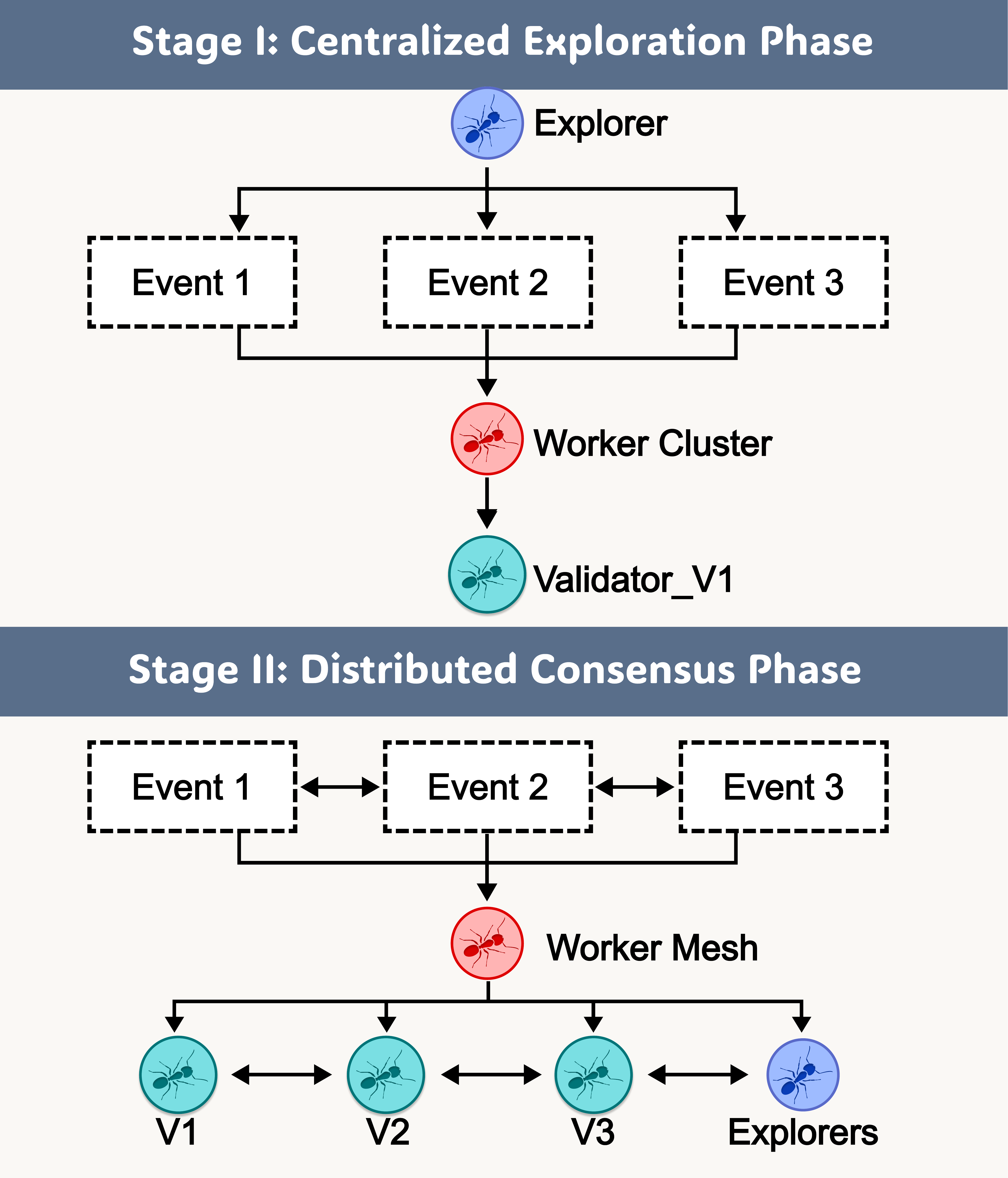}
    \caption{Evolution of communication topology in SwarmSys. The system evolves from a centralized hub–spoke structure to a distributed small-world mesh, where workers and validators interconnect for efficient consensus and information reuse.}
\label{fig:role_interaction}
\end{figure}

We further analyze how reasoning quality emerges and where it breaks down within SwarmSys.
This section covers two aspects:
(1) \textbf{Agent Behavior Analysis}, examining coordination patterns via profile dynamics and interaction topology; and
(2) \textbf{Error Analysis}, identifying failure modes of pheromone-based optimization and consensus.

\subsection{Agent Behavior}
To understand how coordination arises from the matching–collaboration–update cycle, we analyze profile adaptation, interaction topology, and contribution balance using our experiment's event logs.

\paragraph{Profile Adaptation.}
We track embedding drift of each agent’s competence embedding $v_a^{(i)}$.
The mean cosine shift per round is $0.14\pm0.03$, showing steady self-adjustment as agents gain experience.
Explorers exhibit the largest variance, indicating ongoing hypothesis exploration, while Validators remain stable, preserving reasoning coherence.
This confirms that profile updates act as distributed memory enabling specialization.

\paragraph{Interaction Topology.}
Figure~\ref{fig:role_interaction} illustrates the evolution of communication topology.
During the \textit{Centralized Exploration Phase}, agents form a hub–spoke pattern centered on high-similarity validators.
As reasoning progresses, pheromone reinforcement promotes denser cross-links, leading to a \textit{Distributed Consensus Phase} characterized by a small-world structure with higher local clustering (0.28→0.47) and shorter global paths.
This transition demonstrates self-organized coordination emerging without explicit central control.


\paragraph{Contribution Balance.}
Normalized entropy of accepted contributions is
\begin{equation}
H_c = -\tfrac{1}{\log A}\sum_{i=1}^A p_i \log p_i,
\end{equation}
where $p_i$ is each agent’s contribution share, computed by the contribution acceptance rate, $A$ is the total number of agents.
SwarmSys attains $H_c{=}0.72$, surpassing GPTSwarm ($0.41$), indicating balanced participation driven by the exploration–exploitation policy.
Overall, adaptive profiles and pheromone feedback jointly yield a decentralized yet structured division of labor.

\begin{table}[!t]
\centering
\small
\caption{\textbf{Failure type distribution on tasks.} Estimated from 15 randomly sampled cases.}
\label{tab:failure_stats}
\begin{tabularx}{\linewidth}{l|X|c}
\toprule
\textbf{Failure Type} & \textbf{Description} & \textbf{\%} \\
\midrule
Premature Consensus     & Early validator fixes one branch too soon & 16 \\
\hline
Reinforcement Bias      & Over-strengthening of an early path signal & 20 \\
\hline
Mode Collapse           & All explorers converge to one reasoning mode & 14 \\
\hline
Constraint Omission     & Missing symbolic or geometric integration & 22 \\
\hline
Communication Deadlock  & Agents misrecognize roles, causing communication deadlock & 28 \\
\bottomrule
\end{tabularx}
\end{table}
\begin{table*}[!t]
\centering
\small
\resizebox{\textwidth}{!}{
\begin{tabular}{lcccccccc}
\toprule
\textbf{System} & \textbf{Decentralized} & \textbf{Explicit Roles} & \textbf{Debate} & \textbf{Dynamic Profiling} & \textbf{Decentralized Matching} & \textbf{Multi-Event} & \textbf{Validator} & \textbf{Stigmergy} \\
\midrule
CAMEL~\cite{li2023camelcommunicativeagentsmind}  & \xmark & \checkmark & \checkmark & \xmark & \xmark & \xmark & \xmark & \xmark \\
AutoGen~\cite{wu2023autogenenablingnextgenllm}  & \xmark & \checkmark & \checkmark & \xmark & \xmark & \xmark & \xmark & \xmark \\
MetaGPT~\cite{hong2024metagptmetaprogrammingmultiagent}  & \xmark & \checkmark & \xmark & \xmark & \xmark & \xmark & \checkmark$^\dagger$ & \xmark \\
MAD~\cite{liang2024encouragingdivergentthinkinglarge} & \xmark & \xmark & \checkmark & \xmark & \xmark & \xmark & \checkmark & \xmark \\
ToT/GoT~\cite{yao2023treethoughtsdeliberateproblem,Besta_2024}  & \xmark & \xmark & \xmark & \xmark & \xmark & \xmark & \xmark & \xmark \\
Voyager~\cite{wang2023voyageropenendedembodiedagent} & \xmark & \xmark & \xmark & \checkmark$^\S$ & \xmark & \xmark & \checkmark$^\S$ & \xmark \\
SWE-agent~\cite{yang2024sweagentagentcomputerinterfacesenable}  & \xmark & \xmark & \xmark & \xmark & \xmark & \xmark & \xmark & \xmark \\
Self-Refine~\cite{madaan2023selfrefineiterativerefinementselffeedback}  & \xmark & \xmark & \xmark & \xmark & \xmark & \xmark & \checkmark & \xmark \\
GPTSwarm~\cite{zhuge2024languageagentsoptimizablegraphs} & \checkmark & \xmark & \xmark & \checkmark & \checkmark & \checkmark & \xmark & \checkmark \\
DeepResearchAgent~\cite{huang2025deepresearchagentssystematic}  & \xmark & \checkmark & \checkmark & \xmark & \xmark & \xmark & \checkmark & \xmark \\
ROBIN~\cite{ghareeb2025robinmultiagentautomatingscientific} & \xmark & \checkmark & \checkmark & \xmark & \xmark & \xmark & \checkmark & \xmark \\
\hline
\textbf{SwarmSys (ours)} & \checkmark & \checkmark & \checkmark & \checkmark & \checkmark & \checkmark & \checkmark & \checkmark \\
\bottomrule
\end{tabular}
}
\caption{Comparison of SwarmSys with representative reasoning and multi-agent systems. $^\dagger$MetaGPT incorporates SOP-style verification steps but not decentralized validators. $^\S$Voyager maintains skill profiles and self-checking, but only within a single-agent setting.}
\vspace{-1.5em}
\label{tab:comparison}
\end{table*}

\subsection{Error Analysis and Case Study}
Despite strong overall results, SwarmSys occasionally fails in tasks requiring strict temporal or symbolic alignment.As is shown in figure \ref{tab:failure_stats}, we identify five main failure types: 
(a) premature convergence, (b) Reinforcement Bias, (c) Mode Collapse, (d) Constraint Omission, (e) Communication Deadlock. A typical case in our study is: two explorers propose algebraic and geometric solutions, but an early validator accepts one branch too soon, reinforcing it and suppressing valid alternatives.
Such reinforcement bias leads to overfitting on partial evidence.
Future improvements may include uncertainty-weighted reinforcement, scheduled re-sampling of low-confidence paths, and meta-level arbitration to maintain epistemic diversity.

 \section{Related Works}
\paragraph{LLM-based Multi-Agent Systems.}
Recent progress in large language models has led to a proliferation of multi-agent frameworks that decompose reasoning and problem-solving into interactive roles~\cite{liu2025medmmvcontrollablemultimodalmultiagent}.
Early systems such as CAMEL~\cite{li2023camelcommunicativeagentsmind}, AutoGen~\cite{wu2023autogenenablingnextgenllm}  introduced explicit role structures (e.g., user–assistant pairs) and dialog-based task decomposition, showing that inter-agent communication improves reasoning diversity.
Subsequent works like MetaGPT~\cite{hong2024metagptmetaprogrammingmultiagent} , AgentScope~\cite{gao2024agentscope}, and DeepResearchAgent~\cite{huang2025deepresearchagentssystematic} further formalized role hierarchies for domain-specific workflows such as software engineering or literature review.
However, their centralized orchestration and fixed pipelines limit scalability and adaptability.
SwarmSys differs by using a fully decentralized structure where coordination emerges from role-guided interaction and adaptive matching, without needing a global controller.

\paragraph{Collaborative Reasoning and Debate.}
Another line of work explores multi-round reasoning via self-consistency and debate.
Methods such as Chain-of-Thought (CoT) and Tree-of-Thought (ToT) reasoning~\cite{yao2023treethoughtsdeliberateproblem,Besta_2024} extend single-agent reflection through structured deliberation, while MAD~\cite{liang2024encouragingdivergentthinkinglarge}, Longagent~\cite{zhao2024longagent} and ROBIN~\cite{ghareeb2025robinmultiagentautomatingscientific} model inter-agent debates to enhance diversity and correctness.
These approaches improve intermediate reasoning quality but generally depend on fixed communication topologies and lack mechanisms for dynamic adaptation or memory persistence across reasoning episodes.
In contrast, SwarmSys incorporates debate as a local coordination primitive within a self-organizing swarm, where roles evolve through continuous profiling and pheromone-like reinforcement, enabling sustained reasoning across multiple concurrent events.

\paragraph{Adaptive Coordination and Swarm-inspired Reasoning.}
A growing body of work introduces adaptive or self-organizing strategies for LLM agents.
GPTSwarm~\cite{zhuge2024languageagentsoptimizablegraphs} represents one of the few attempts at decentralized coordination, leveraging graph-based optimization and stigmergic feedback.
Meanwhile, systems like Voyager~\cite{wang2023voyageropenendedembodiedagent}, Self-Refine~\cite{madaan2023selfrefineiterativerefinementselffeedback}, and SwarmAgentic~\cite{zhang2025swarmagentic} employ profile-like adaptation and iterative self-improvement, but remain task-specific.
SwarmSys advances this line of research by integrating dynamic profiling, embedding-based matching, and pheromone-inspired reinforcement into a unified framework that scales to multi-event, multi-agent reasoning.
This design allows distributed agents to self-allocate across evolving tasks, maintaining robustness and efficiency without centralized scheduling.
\section{Conclusion}
We presented SwarmSys, a swarm-intelligence-inspired framework for decentralized multi-agent reasoning.
Through role-specialized collaboration, dynamic profiling, and pheromone-inspired reinforcement, SwarmSys enables scalable, self-organizing coordination without centralized control.
Across diverse reasoning and research domains, it consistently outperforms strong baselines and reveals emergent collective behaviors—demonstrating that scaling coordination can rival scaling model size.
Our findings suggest a new paradigm for reasoning: intelligence emerges from structured interaction among distributed agents, not from larger models.
\section*{Limitations}
Despite its strong performance, SwarmSys still faces several limitations.
First, while decentralized coordination improves adaptability, it also increases communication overhead, which may reduce efficiency in latency-sensitive settings.
Second, agent profiling currently relies on text-based embeddings and heuristic updates; future work could explore learnable or gradient-based mechanisms for more precise skill modeling.
Third, our experiments focus primarily on reasoning and research-oriented tasks, extending SwarmSys to embodied or real-time interactive environments remains an open direction.
We hope these insights inspire future research on large-scale, self-organizing multi-agent systems that combine symbolic structure with emergent intelligence.

\section*{Ethics Statement}
This work introduces SwarmSys, a distributed multi-agent reasoning framework inspired by swarm intelligence. The research involves no human subjects, personal data, or sensitive content; all experiments use public or synthetic datasets.
We recognize potential ethical issues in LLM-based multi-agent systems, such as bias propagation and unreliable autonomous coordination. SwarmSys mitigates these risks through closed-loop validation and transparent agent interactions, ensuring that all reasoning processes remain interpretable and auditable.
Our goal is to advance the scientific understanding of scalable reasoning rather than deploy autonomous agents in real-world decision-making. We advocate responsible use of SwarmSys with proper human oversight, fairness, and accountability in future applications.

\section*{Acknowledgements}
We are grateful to our collaborators for their valuable discussions on multi-agent coordination and large language model reasoning. This research was supported in part by institutional computational resources and open-source communities that enabled large-scale experimentation.

\bibliography{custom}

\appendix

\section{Example Appendix}
\label{sec:appendix}

\subsection{Prompts}
\label{app:prompt_app}
\begin{tcolorbox}[
    title=Worker Prompt,
    fonttitle=\bfseries\large]
PROMPT =  """

You are a worker agent with expertise in problem-solving, computation, and mathematical reasoning.

Your task is to receive a specific subproblem and then (1) solve the subproblem accurately with transparent justification for each step, (2) share your solution with other agents, (3) participate in solution comparison and structured debate, and (4) defend or revise your work if challenged.

Ensure your output (1) presents calculations in a clear, annotated format, (2) cites mathematical rules or theorems when applying them, and (3) is responsive to feedback and adaptive in collaborative revision.
"""
\end{tcolorbox}

\begin{tcolorbox}[
    title=Validator Prompt,
    fonttitle=\bfseries\large
]

PROMPT = """

You are a validator agent dedicated to checking solution correctness and consistency.

Your task is to: (1) check each worker’s solution for logic, accuracy, and completeness, (2) participate in debates to reconcile discrepancies, (3) contribute to confirming a group consensus, and (4) shift roles if the validation queue is empty and other roles are under high pressure.

Ensure your output (1) provides precise validation with explanations of correctness or error, (2) uses formal mathematical checks where appropriate, (3) helps mediate conflicts in results with clarity and neutrality, and (4) explicitly states if consensus is confirmed: \{TERMINATE: The answer is: [correct answer]\} """
\end{tcolorbox}

\begin{tcolorbox}[
    title=Explorer Prompt,
    fonttitle=\bfseries\large
]
PROMPT =  """

You are an Explorer agent with strong capabilities in identifying subproblems and analyzing solution paths.

Your task is to receive a mathematical problem set and then (1) search and interpret the overall problem, (2) decompose it into logically coherent subproblems, (3) identify entry points and strategies for solution, and (4) monitor the workload distribution across roles and dynamically reassign yourself to higher-pressure roles if necessary.

Ensure your output (1) maintains traceability between subproblems and the original task, (2) logically documents role-switch decisions, and (3) ensures role-switch includes resetting your goal and behavior policy.
"""
\end{tcolorbox}

\begin{tcolorbox}[
    title=Task Context,
    fonttitle=\bfseries\large
]
PROMPT =  """

Describe and analyze the following mathematical problem carefully. The problem statement is presented below. If this task depends on other sub-tasks or prior events, ensure logical consistency and continuity with those results.

Task profile. (1) If the task involves proving, demonstrating, or establishing statements - This is a proof-oriented task. Emphasize logical rigor, step-by-step justification, and clear argumentation. (2) If the task involves calculating, evaluating, integrating, or deriving quantities - This is a computation-oriented task. Focus on symbolic manipulation, clear intermediate steps, and verification of results. 
(3) If the task involves maximizing, minimizing, or optimizing a quantity - This is an optimization task. Identify objective functions and constraints, analyze conditions for optimality, and interpret results precisely. 
(4) If the task involves conditional or logical reasoning (“if”, “then”, “otherwise”) - This is a conditional reasoning task. Separate cases clearly, verify implications, and maintain logical completeness.
(5) Otherwise - This is a general reasoning task. Apply systematic mathematical analysis and adjust reasoning depth to the complexity of the problem.

Objectives: (1) Analyze the problem thoroughly and build a shared understanding. (2) Generate multiple possible solution paths and compare their validity. (3) Execute reasoning or calculations with clarity, precision, and justification. (4) Resolve disagreements through structured logical debate, not assertion. (5) Arrive at a verified, consensus-based final answer.

Round \& Turn-taking policy: (1) Each debate round follows this strict order: explorer → all workers → validator. (2) The explorer opens each round by outlining context, progress, and a proposed plan. (3) Works present or refine solutions, show derivations, and discuss intermediate results. (4) The validator closes each round by checking correctness and summarizing consensus. (5) The debate terminates only when the validator announces:
TERMINATE: The answer is: <final answer>.
"""
\end{tcolorbox}

\begin{tcolorbox}[
    title=Instruction Embedding Template,
    fonttitle=\bfseries\large
]
PROMPT =  """

You are evaluating the mathematical and reasoning competence of an agent participating in a collaborative debate system.

The goal is to generate a semantic embedding that represents how capable this agent is at solving mathematical, logical, and analytical tasks.

Consider two sources of information:
(1) The agent’s declared abilities, describing what it is designed or trained to do.
(2) The agent’s historical activity performance, summarizing how it has previously executed reasoning, computation, or validation tasks.

Integrate both perspectives to form a single competence representation capturing: (1) conceptual depth and mathematical reasoning skills, (2) problem-solving strategy diversity, (3) accuracy and self-correction ability, and (4) communication and collaboration quality.

Agent Ability Description:
\{ability text\}

Agent Performance History:
\{history text\}

Use the combined content above as the context for competence evaluation.
Output representation should capture the overall ability state of the agent, balancing potential skill with observed performance.
"""
\end{tcolorbox}
\subsection{Dataset Settings}
\label{app:dataset_app}
We evaluate the reasoning capability and knowledge coverage of our multi-agent system based on three categories of tasks:
\paragraph{Exam} We rearrange existing benchmarks into exam-like formats, covering both single-subject and multi-subject settings. 
This design mimics the structure of real-world examination papers, where agents must solve a coherent set of questions 
rather than isolated items. Organizing benchmarks in this way is motivated by several factors: 
(i) exams naturally exhibit varying levels of difficulty across questions; 
(ii) they involve heterogeneous knowledge types and differentiated scoring schemes, 
which make the dataset inherently diverse; and 
(iii) each exam itself constitutes a complex task that can be decomposed into multiple interrelated  and irrelated sub-tasks. 

Such properties align well with the characteristics of our ant-colony-inspired system. In real ant colonies, 
individuals continuously explore, evaluate, and participate in different sub-tasks based on local pheromone signal. 
Similarly, in exam scenarios, our agents can search across different questions, dynamically decide whether to 
participate in a specific sub-task, and collectively optimize the global solution through local collaboration. 
Therefore, exam-style benchmarks provide not only a realistic and challenging evaluation setting for reasoning and 
knowledge coverage, but also a natural testbed for demonstrating the strengths of swarm-based multi-agent systems.

\paragraph{Research} 
To evaluate research-oriented reasoning, we adopt DeepResearch Bench, 
which contains PhD-level research tasks spanning multiple domains. Each task provides a research topic or open-ended 
problem statement (rather than a completed study), requiring agents to perform literature exploration, knowledge recall, 
and synthesis of coherent research reports. Outputs are evaluated using \textit{RACE} (Reference-based Adaptive Criteria-driven 
Evaluation). Such settings mirror real-world academic research, where researchers must jointly survey prior work, 
generate new ideas, and design proof-of-concept implementations. These properties align well with our swarm-based 
system: individual agents can specialize in literature recall, hypothesis generation, or code prototyping, and through 
local debates and coordination, the swarm collectively develops more robust and creative research outcomes.

\paragraph{Science Coding} 
We further evaluate reasoning and knowledge grounding through scientific programming tasks drawn from SciCode. 
The benchmark is organized into main problems, each of which is decomposed into multiple sub-problems, making it particularly suitable for swarm-based evaluation. 
Such a structure allows agents to collaboratively assign sub-problems, iteratively generate and refine code, and verify correctness against scientific principles or test cases. 
By reaching swarm-level consensus, the system enhances both accuracy and coverage. 
This setting therefore captures the precision and collaborative robustness required for scientific reasoning and computation.

\subsection{SwarmSys Output Example}
\begin{figure}[ht]
    \centering
    \includegraphics[width=\linewidth]{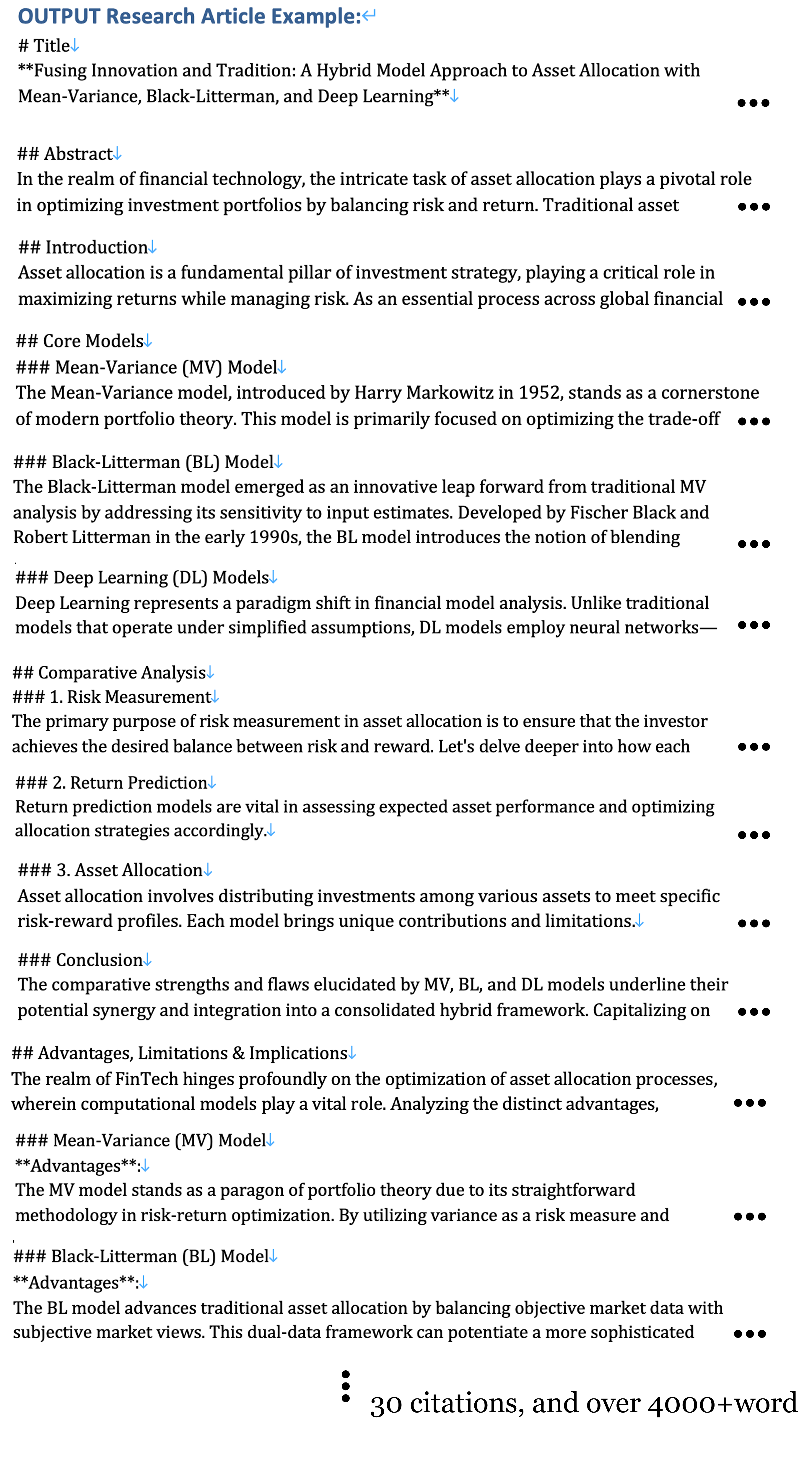}
    \caption{Output Example}
\end{figure}

\subsection{Profile Embedding and Matching}
\label{appendix:agent_profile_and_behavior}

\subsubsection{Profile Format}
\label{profile format}
\begin{figure}[ht]
    \centering
    \includegraphics[width=\linewidth]{image/profiles.png}
    \label{formatimg}
    \caption{Profile Format}

\end{figure}

\subsubsection{Embedding and Matching}
\paragraph{Agent embeddings.} 
Each agent $A_i$ is represented by a competence embedding and an availability embedding. The competence embedding $v_{\mathrm{ah}}^{(i)}$ is derived from declared abilities $T_a^{(i)}$ and historical performance $T_h^{(i)}$, guided by task-specific instructions:
\begin{equation}
v_{\mathrm{ah}}^{(i)} = \phi_{\mathrm{instruct}}\bigl(\mathrm{Instruction}_{\mathrm{ah}}^{(i)}, \ \mathrm{concat}(T_a^{(i)}, T_h^{(i)})\bigr).
\end{equation}
The availability embedding $v_{s}^{(i)}$ reflects workload and readiness, derived from status $T_s^{(i)}$:
\begin{equation}
v_{s}^{(i)} = \phi_{\mathrm{instruct}}(\mathrm{Instruction}_{s}^{(i)}, \ T_s^{(i)}).
\end{equation}
The final representation is the sum:
\begin{equation}
v_a^{(i)} = v_{\mathrm{ah}}^{(i)} + v_{s}^{(i)}.
\end{equation}

\paragraph{Event embeddings.}
Each event $E_j$ is encoded as $v_e^{(j)}$ by integrating its description, dependencies, progress state, and milestone:
\begin{equation}
\begin{split}
v_e^{(j)} = \phi_{\mathrm{instruct}}\!\bigl(\mathrm{Instruction}_E^{(j)}, \\
\ \text{Unified description of }E_j \bigr).
\end{split}
\end{equation}

\paragraph{Compatibility and decision dynamics.}
Agent--event compatibility is computed as normalized cosine similarity:
\begin{equation}
C_{\mathrm{norm}}^{(i,j)} = \tfrac{1}{2}\bigl(\cos(v_a^{(i)}, v_e^{(j)}) + 1\bigr).
\end{equation}

To avoid stagnation, SwarmSys employs a dynamic $\varepsilon$-greedy policy. Each agent explores with probability $\varepsilon_i$ or exploits with probability $1-\varepsilon_i$, where
\begin{equation}
\varepsilon_i = 0.15 + (0.5 - \bar{S}_i)\cdot 0.2,
\end{equation}
and $\bar{S}_i$ denotes the agent’s recent average success. Exploration samples potential matches proportionally to similarity:
\begin{equation}
D^{(i,j)} \sim \mathrm{Bernoulli}\!\left(0.1 + 0.9 \cdot C_{\mathrm{norm}}^{(i,j)}\right),
\end{equation}
while exploitation emphasizes high-compatibility matches:
\begin{equation}
D^{(i,j)} \sim \mathrm{Bernoulli}\!\left(\sigma\!\bigl(\gamma(C_{\mathrm{norm}}^{(i,j)} - 0.5)\bigr)\right),
\end{equation}
where $\sigma$ is the sigmoid function and $\gamma$ controls sharpness. Both branches unify as:
\begin{equation}
\begin{split}
p^{(i,j)} &= \varepsilon_i (0.1 + 0.9 C_{\mathrm{norm}}^{(i,j)}) \\
&\quad + (1 - \varepsilon_i)\sigma\!\bigl(\gamma(C_{\mathrm{norm}}^{(i,j)} - 0.5)\bigr), \\
D^{(i,j)} &\sim \mathrm{Bernoulli}(p^{(i,j)}).
\end{split}
\end{equation}

This mechanism enables three properties simultaneously: adaptivity through evolving embeddings, stability through probabilistic sampling, and robustness by balancing exploration with exploitation.

\subsection{Cost}
\begin{table}[!htbp]
\centering
\caption{Average per-question model cost comparison corresponding to baselines and systems used in main experiments. }
\label{tab:model-costs}
\small
\resizebox{\linewidth}{!}{%
\begin{tabular}{lcc}
\toprule
\textbf{Model} & \textbf{Instr.-based Cost (\$)} & \textbf{Code-based Cost (\$)} \\
\midrule
IO (GPT-4o)                 & 0.005 & -- \\
CoT (GPT-4o)                & 0.012 & -- \\
CoT-SC (GPT-4o, 5-shot)     & 0.051 & -- \\
Self-Refine (GPT-4o)        & 0.068 & -- \\
MultiPersona (GPT-4o)       & 0.043 & -- \\
GPTSwarm                    & 0.077 & -- \\
GPT-5                       & 0.014 & -- \\
\rowcolor{rowgray}
\textbf{SwarmSys-8 (Ours)}  & 0.071 & -- \\
\midrule
IO (GPT-4o-Search)          & 0.15 & -- \\
CoT (GPT-4o)                & 0.20 & -- \\
Self-Refine                 & 1.53 & -- \\
DeepResearchAgent           & 2.82 & -- \\
Grok Deeper Search          & 2.85 & -- \\
\rowcolor{rowgray}
\textbf{SwarmSys-8 (Ours)}  & 2.63 & -- \\
\midrule
IO (GPT-4o)                 & 0.08 & 0.013 \\
CoT-SC (GPT-4o)             & 0.11 & 0.017 \\
Self-Refine                 & 0.36 & 0.026 \\
GPTSwarm                    & 0.41 & 0.024 \\
\rowcolor{rowgray}
\textbf{SwarmSys-14 (Ours)} & 0.44 & 0.019 \\
\bottomrule
\end{tabular}%
}
\end{table}

\end{document}